\documentclass[10pt,twocolumn,letterpaper]{article}

\usepackage{cvpr}
\usepackage{times}
\usepackage{epsfig}
\usepackage{graphicx}
\usepackage{amsmath}
\usepackage{amssymb}
\usepackage{multirow}
\usepackage{multicol}
\usepackage{mathtools}
\usepackage{bbm}
\usepackage{enumitem}
\usepackage{subfig}
\usepackage{authblk}
\usepackage{floatrow}
\usepackage[marginal]{footmisc}
\usepackage[breaklinks=true,bookmarks=false]{hyperref}

\newfloatcommand{capbtabbox}{table}[][\FBwidth]

\newtheorem{myThm}{Theorem}
\newtheorem{prf}{Proof}

\def \Pset {\mathcal{P}}

\newcommand{\mypar}[1]{{\bf #1.}}

\cvprfinalcopy 

\def\cvprPaperID{1911} 

\begin{document}

\title{Actional-Structural Graph Convolutional Networks for \\ Skeleton-based Action Recognition}

\author[1]{Maosen Li}
\author[2]{Siheng Chen}
\author[1]{Xu Chen}
\author[1]{Ya Zhang}
\author[1]{Yanfeng Wang}
\author[3]{Qi Tian}

\affil[1]{{ Cooperative Medianet Innovation Center, Shanghai Jiao Tong University}}
\affil[2]{{ Mitsubishi Electric Research Laboratories}}
\affil[3]{{ Huawei Noah's Ark Lab}
\authorcr {\tt\small \{maosen\_li, xuchen2016, ya\_zhang, wangyanfeng\} @sjtu.edu.cn, sihengc@andrew.cmu.edu, tian.qi1@huawei.com}}

\maketitle
\pagestyle{empty}
\thispagestyle{empty}

\begin{abstract}
   Action recognition with skeleton data has recently attracted much attention in computer vision. Previous studies are mostly based on fixed skeleton graphs, only capturing local physical dependencies among joints, which may miss implicit joint correlations. To capture richer dependencies, we introduce an encoder-decoder structure, called A-link inference module, to capture action-specific latent dependencies, i.e. actional links, directly from actions. We also extend the existing skeleton graphs to represent higher-order dependencies, i.e. structural links. Combing the two types of links into a generalized skeleton graph, we further propose the actional-structural graph convolution network (AS-GCN), which stacks actional-structural graph convolution and temporal convolution as a basic building block, to learn both spatial and temporal features for action recognition. A future pose prediction head is added in parallel to the recognition head to help capture more detailed action patterns through self-supervision. We validate AS-GCN in action recognition using two skeleton data sets, NTU-RGB+D and Kinetics. The proposed AS-GCN achieves consistently large improvement compared to the state-of-the-art methods. As a side product, AS-GCN also shows promising results for future pose prediction. Our code is available at \url{https://github.com/limaosen0/AS-GCN}. \footnote{Accepted by CVPR 2019.}
\vspace{-10pt}
\end{abstract}

\section{Introduction}
Human action recognition, broadly applicable to video surveillance, human-machine interaction, and virtual reality~\cite{6126548, 1032808, Sudha2017Approaches}, has recently attracted much attention in computer vision. Skeleton data, representing dynamic 3D joint positions, have been shown to be effective in action representation, robust against sensor noise, and efficient in computation and storage \cite{ijcai_ChaoLi, Yan2017}. The skeleton data are usually obtained by either locating 2D or 3D spatial coordinates of joints with depth sensors or using pose estimation algorithms based on videos~\cite{Cao_2017_CVPR}.

\begin{figure}
\centering
\includegraphics[width=8.2cm]{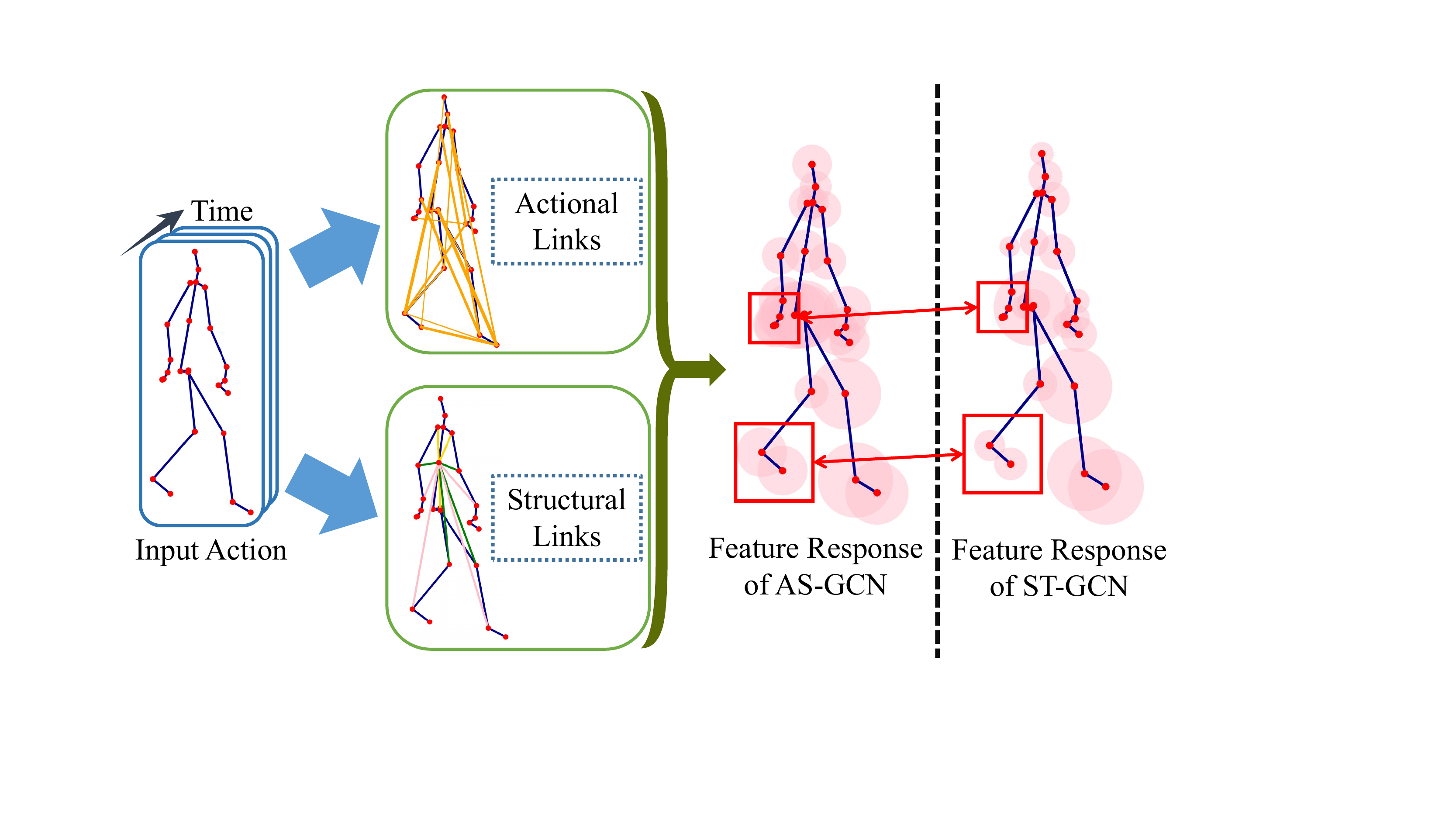} 
\caption{\small Feature learning with generalized skeleton graphs. the actional links and structural links capture dependencies between joints. For the action ``walking'', actional links denotes that hands and feet are correlated. The semilucent circles on the right bodies are the joint feature maps for recognition, whose areas are the response magnitudes. Compared to ST-GCN, AS-GCN obtains responses on collaborative moving joints (red boxes).}
\label{fig:function}
\vspace{-5pt}
\end{figure}


\begin{figure*}
\centering
\includegraphics[width=14cm]{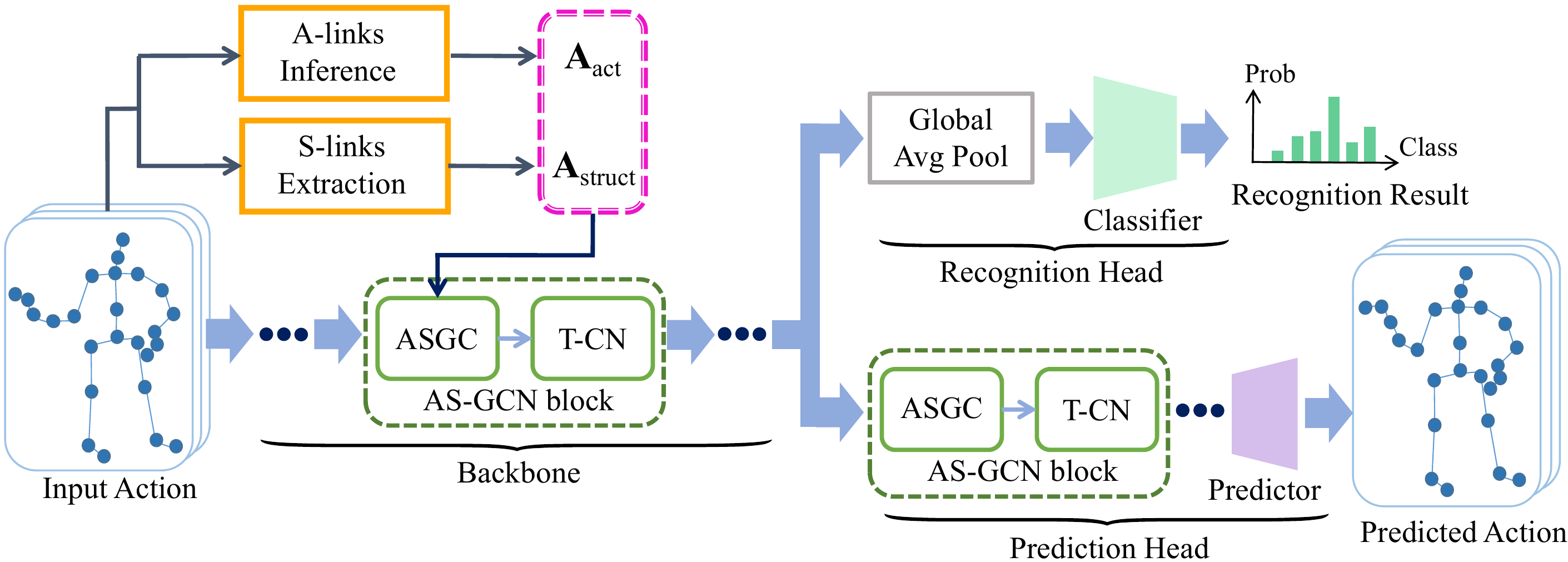} 
\caption{\small The pipeline of the proposed AS-GCN. The inferred actional graph \emph{A-links} and extended structural graph \emph{S-links} are fed to the AS-GCN blocks to learn spatial features. The last AS-SCN block is connect to two parallel branches, the recognition head and the prediction head, which are simultaneously trained.}
\label{fig:pipeline} 
\end{figure*}
The earliest attempts of skeleton action recognition often encode all the body joint positions in each frame to a feature vector for pattern learning~\cite{Vemulapalli_2014_CVPR, Fernando_2015_CVPR, Du_2015_CVPR, vis_cnn}. These models rarely explore the internal dependencies between body joints, resulting to miss abundant actional information. To capture joint dependencies, recent methods construct a skeleton graph whose vertices are joints and edges are bones, and apply graph convolutional networks (GCN) to extract correlated features~\cite{kipf_iclr2017}. The spatio-temporal GCN (ST-GCN) is further developed to simultaneously learn spatial and temporal features~\cite{AAAI1817135}. ST-GCN though extracts the features of joints directly connected via bones, structurally distant joints, which may cover key patterns of actions, are largely ignored. For example, while walking, hands and feet are strongly correlated. While ST-GCN tries to aggregate wider-range features with hierarchical GCNs, node features might be weaken during long diffusion \cite{AAAI1816098}. 

We here attempt to capture richer dependencies among joints by constructing generalized skeleton graphs. In particular, we data-driven infer the actional links (A-links) to capture the latent dependencies between any joints. Similar to~\cite{pmlr-v80-kipf18a}, an A-link inference module (AIM) with an encoder-decoder structure is proposed. We also extend the skeleton graphs to represent higher order relationships as the structural links (S-links). Based on the generalized graphs with the A-links and S-links, we propose an actional-structural graph convolution to capture spatial features. We further propose the actional-structural graph convolution network (AS-GCN), which stacks multiple of actional-structural graph convolutions and temporal convolutions. As a backbone network, AS-GCN adapts various tasks. Here we consider action recognition as the main task and future pose prediction as the side one. The prediction head promotes self-supervision and improve recognition by preserving detailed features. Figure~\ref{fig:function} presents the characteristics of the AS-GCN model, where we learn the actional links and extend the structural links for action recognition. The feature responses present that we could capture more global joint information than ST-GCN, which only uses the skeleton graph to model the local relations.

To verify the effectiveness of the proposed AS-GCN, we conduct extensive experiments on two distinct large-scale data sets: NTU-RGB+D \cite{Shahroudy_2016_CVPR} and Kinetics \cite{DBLP:journals/corr/KayCSZHVVGBNSZ17}. The experiments have demonstrated that AS-GCN outperforms the state-of-the-art approaches in action recognition. Besides, AS-GCN accurately predicts future frames, showing that sufficient detailed information is captured. The main contributions in this paper are summarized as follows:

\begin{itemize}[nosep, wide=0pt, leftmargin=*, after=\strut]
\item 
We propose the A-link inference module (AIM) to infer actional links which capture action-specific latent dependencies. The actional links are combined with structural links as generalized skeleton graphs; see Figure~\ref{fig:function};
\item
We propose the actional-structural graph convolution network (AS-GCN) to extract useful spatial and temporal information based on the multiple graphs; see Figure~\ref{fig:pipeline};
\item
We introduce an additional future pose prediction head to predict future poses, which also improves the recognition performance by capturing more detailed action patterns;
\item
The AS-GCN outperforms several state-of-the-art methods on two large-scale data sets; As a side product, AS-GCN is also able to precisely predict the future poses.
\end{itemize}

\section{Related Works}
Skeleton data is widely used in action recognition. Numerous algorithms are developed based on two approaches: the hand-crafted-based and the deep-learning-based. The first approach designs algorithms to capture action patterns based on the physical intuitions, such as local occupancy features~\cite{cvpr_wang}, temporal joint covariances~\cite{ijcai_Hussein} and Lie group curves~\cite{Vemulapalli_2014_CVPR}. On the other hand, the deep-learning-based approach automatically learns the action faetures from data. Some recurrent-neural-network (RNN)-based models capture the temporal dependencies between consecutive frames, such as bi-RNNs~\cite{Du_2015_CVPR}, deep LSTMs~\cite{Shahroudy_2016_CVPR, 10.1007/978-3-319-46487-9_50}, and attention-based model~\cite{AAAI1714437}.  Convolutional neural networks (CNN) also achieve remarkable results, such as residual temporal CNN~\cite{a8014941}, information enhancement model~\cite{vis_cnn} and CNN on action representations~\cite{Ke_2017_CVPR}. Recently, with the flexibility to exploit the body joint relations, the graph-based approach draws much attention~\cite{AAAI1817135, Si_2018_ECCV}. In this work, we adopt the graph-based approach for action recognition. Different from any previous method, we learn the graphs adaptively from data, which captures useful non-local information about actions.
 
\section{Background}
In this section, we cover the background material necessary for the rest of the paper. 
\subsection{Notations}
We consider a skeleton graph as ${\cal G}(V,E)$, where $V$ is the set of $n$ body joints and $E$ is a set of $m$ bones. Let $\mathbf{A} \in \{0,1\}^{n\times n}$ be the adjacent matrix of the skeleton graph, where $\mathbf{A}_{i,j} = 1$ if the $i$-th and the $j$-th joints are connected and $0$ otherwise. ${\bf A}$ fully describes the skeleton structure. Let $\mathbf{D} \in \mathbb{R}^{n\times n}$ be the diagonal degree matrix, where $\mathbf{D}_{i,i}=\sum_j\mathbf{A}_{i,j}$.  To capture more refined location information, we part one root node and its neighbors into three sets, including 1) the root node itself, 2) the centripetal group, which are closer to the body barycenter than root, and 3) the centrifugal group, and $\mathbf{A}$ is accordingly parted to be $\mathbf{A}^{\rm (root)}$, $\mathbf{A}^{\rm (centripetal)}$ and $\mathbf{A}^{\rm (centrifugal)}$. We denote the partition group set as $\Pset{}=\{\text{root},\text{centripetal}, \text{centrifugal}\}$.
Note that $\sum_{p \in \Pset{}} \mathbf{A}^{(p)} = \mathbf{A}$. Let $\mathcal{X} \in \mathbb{R}^{n \times 3 \times T}$ be the 3D joint positions across $T$ frames. Let $\mathbf{X}_t = \mathcal{X}_{:, :, t} \in \mathbb{R}^{n \times 3}$ be the 3D joint positions at the $t$-th frame, which slices the $t$-th frame in the last dimension of $\mathcal{X}$. Let  $\mathbf{x}_{i}^t  = \mathcal{X}_{i, :, t}  \in  \mathbb{R}^d$ be  the positions of the $i$-th joint at the $t$-th frame.

\subsection{Spatio-Temporal GCN}
Spatio-temporal GCN (ST-GCN)~\cite{AAAI1817135} consists of a series of the ST-GCN blocks. Each block contains a spatial graph convolution followed by a temporal convolution, which alternatingly extracts spatial and temporal features. The last ST-GCN block is connected to a fully-connected classifier to generate final predictions. The key component in ST-GCN is the spatial graph convolution operation, which introduces weighted average of neighboring features for each joint. Let $\mathbf{X}_{\rm{in}}\in\mathbb{R}^{n\times d_{\rm{in}}}$ be the input features of all joints in one frame, where $d_{\rm{in}}$ is the input feature dimension, and $\mathbf{X}_{\rm{out}}\in\mathbb{R}^{n\times d_{\rm{out}}}$ be the output features obtained from spatial graph convolution, where $d_{\rm{out}}$ is the dimension of output features. The spatial graph convolution is
\begin{equation}
    \label{eq:st_gcn_gcn}
  \mathbf{X}_{\rm{out}} =
    \sum_{p \in \Pset{}}
    \mathbf{M}_{\rm{st}}^{(p)} \circ 
	\widetilde{\mathbf{A}^{(p)}}  
   \mathbf{X}_{\rm{in}}
    \mathbf{W}_{\rm{st}}^{(p)},
\end{equation}
where $\widetilde{\mathbf{A}^{(p)}} =  \mathbf{D}^{{(p)}^{-\frac{1}{2}}}\mathbf{A}^{(p)}  \mathbf{D}^{{(p)}^{-\frac{1}{2}}}  \in  \mathbb{R}^{n \times n}$ is the normalized adjacent matrix for each partition group, $\circ$ denotes the Hadamard product, $\mathbf{M}_{\rm{st}}^{(p)}\in \mathbb{R}^{n \times n}$ and $\mathbf{W}_{\rm{st}}^{(p)}\in \mathbb{R}^{n \times d_{\rm{out}}}$ are trainable weights for each partition group to capture edge weights and feature importance, respectively.

\section{Actional-Structural GCN} 
\begin{figure}
\centering
\includegraphics[width=7.7cm]{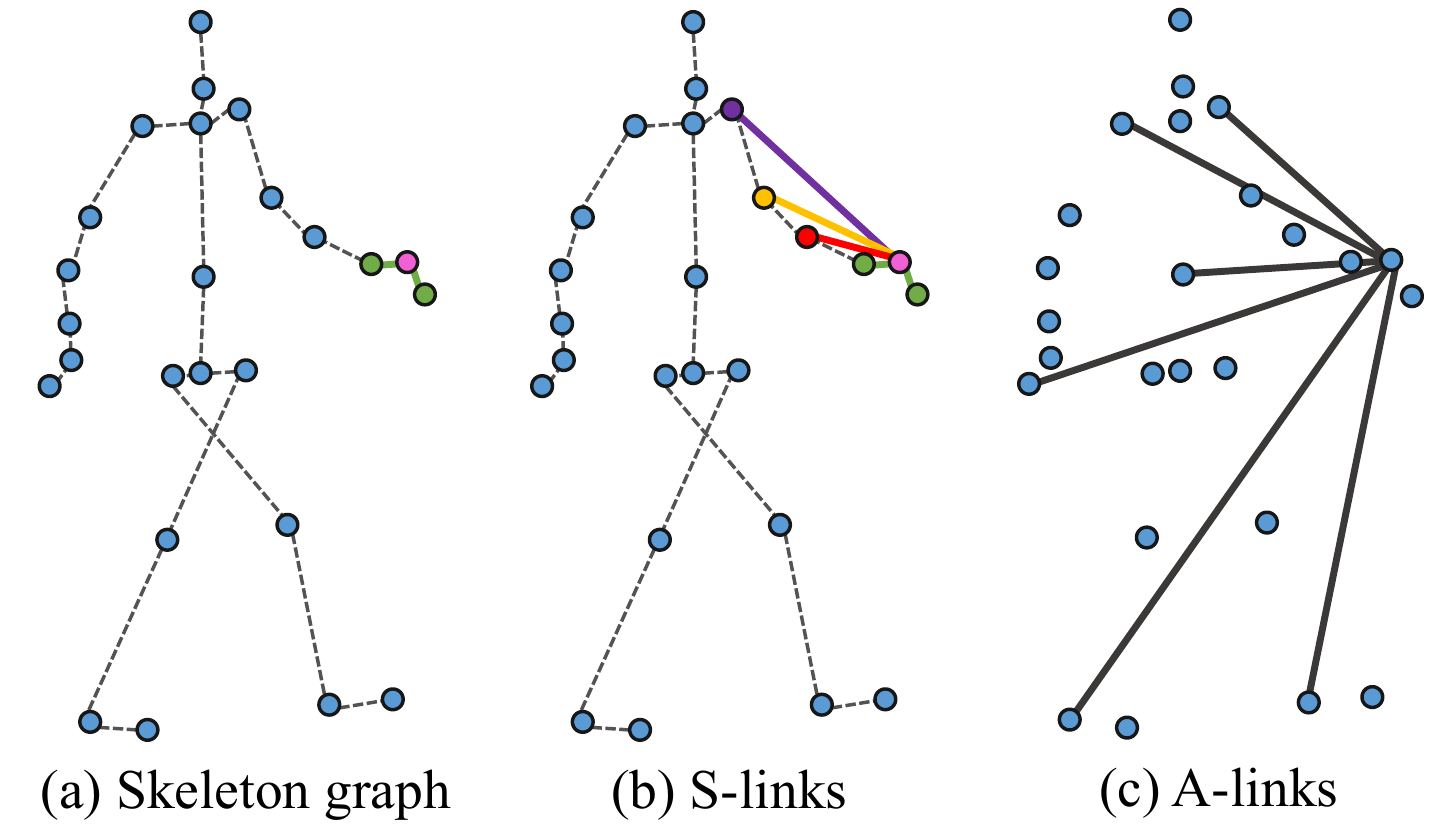} 
\caption{\small An example of the skeleton graph, S-links and A-links for walking. In each plot, the links from "Left Hand" to its neighbors are shown in solid lines. (a) Skeleton links with limited neighboring range; (b) S-links, allowing "Left Hand" to link to the entire arm;  (c) A-links, capturing long-range action-specific relations.}
\label{fig:h-links} 
\end{figure}

The generalized graph, named~\emph{actional-structural graph}, is defined as ${\cal G}_g(V, E_g)$, where $V$ is the original set of joints and  $E_g$ is the set of generalized links. There are two types of links in $E_g$: structural links (S-links), explicitly derived from the body structure, and actional links (A-links), directly inferred from skeleton data. See the illustration of both types in Figure~\ref{fig:h-links}.

\subsection{Actional Links (A-links)} 
\begin{figure}
\centering
\includegraphics[width=8.2cm]{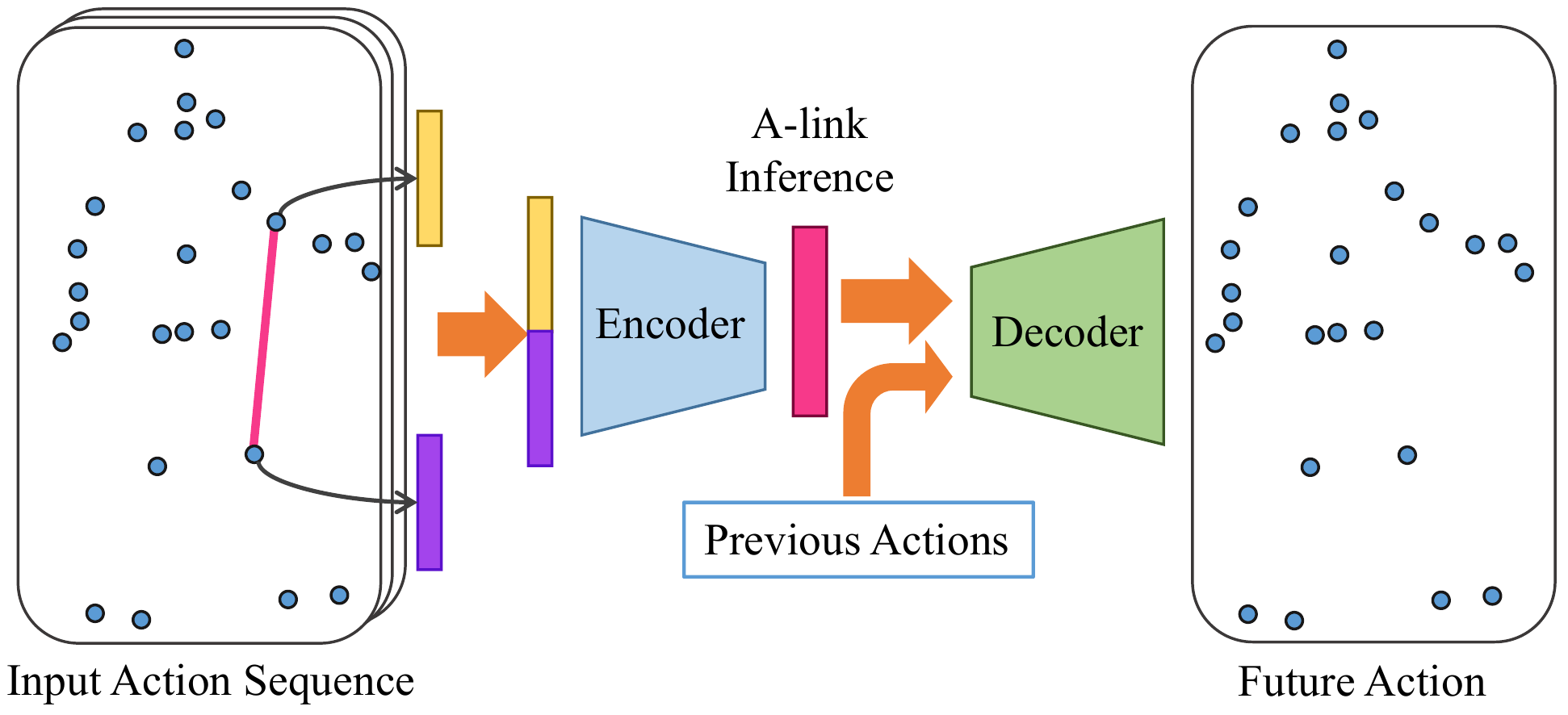} 
\caption{\small A-links inference module (AIM). To infer the A-link between two joints, the joint features are concatenated and fed into the encoder-decoder formed AIM. The encoder produces the inferred A-links and the decoder generates the future pose conditioned on the A-links and previous actions.}
\label{fig:AIM}
\end{figure}

Many human actions need far-apart joints to move collaboratively, leading to non-physical dependencies among joints. To capture corresponding dependencies for various actions, we introduce actional links (A-links), which are activated by actions and might exist between arbitrary pair of joints. To automatically infer the A-links from actions, we develop a trainable~\emph{A-link inference module} (AIM), which consists of an encoder and a decoder. The encoder produces A-links by propagating information between joints and links iteratively to learn link features; and the decoder predict future joint positions based on the inferred A-links; see Figure~\ref{fig:AIM}.
 We use AIM to warm-up the A-links, which are further adjusted during the training process.

\mypar{Encoder} The functionality of an encoder is to estimate the states of the A-links given the 3D joint positions across time; that is,  
\begin{equation}
\label{eq:aim_encode}
 \mathcal{A} \ = \ {\rm encode} ( \mathcal{X}) \in [0,1]^{n \times n \times C},
\end{equation}
where $C$ is the number of A-link types. Each element $\mathcal{A}_{i,j, c}$ denotes the probability that the $i,j$-th joints are connected with the $c$-th type. The basic idea to design the mapping ${\rm encode} (\cdot)$ is to first exact link features from 3D joint positions and then convert the link features to the linking probabilities. To exact link features, we propagate information between joints and links alternatingly. Let $\mathbf{x}_i = {\rm vec} \left(\mathcal{X}_{i, :, :} \right) \in \mathbb{R}^{ dT }$ be the vector representation of the $i$-th joint's feature across all the $T$ frames. We initialize the joint feature $\mathbf{p}_{i}^{(0)} = \mathbf{x}_i$. In the $k$-th iteration, we propagate information back and forth between joints and links,
 \begin{equation*}
     \begin{aligned}
     {\rm link~features:}~\mathbf{Q}_{i, j}^{(k+1)}   &=  f_{e}^{(k)}(f_{v}^{(k)}(\mathbf{p}_{i}^{(k)} )\oplus (f_{v}^{(k)}(\mathbf{p}_{j}^{(k)} )),
\\
    {\rm joint~features:}~\mathbf{p}_{i}^{(k+1)}  &=  \textstyle\mathcal{F}(  \mathbf{Q}_{i,: }^{(k+1)}    )\oplus  \mathbf{p}_{i}^{(k)},
     \end{aligned}
 \end{equation*}
where $f_v(\cdot)$ and $f_e(\cdot)$ are both multi-layer perceptrons, $\oplus$ is vector concatenation, and $\mathcal{F}(\cdot)$ is an operation to aggregate link features and obtain the joint feature; such as averaging and elementwise maximization. After propagating for $K$ times, the encoder outputs the linking probabilities as
\begin{equation}
\label{eq:aim_encode_output}
  \mathcal{A}_{i,j, :} ={\rm softmax}\left( \frac{\mathbf{Q}_{i,j}^{(K)} +\mathbf{r} }{\tau} \right)\in \mathbb{R}^{C},
\end{equation}
where $\mathbf{r}$ is a random vector, whose elements are i.i.d. sampled from $\rm{Gumbel(0,1)}$ distribution and $\tau$ controls the discretization of $\mathcal{A}_{i,j,:}$. Here we set $\tau=0.5$. We obtain the linking probabilities $\mathcal{A}_{i,j, :} $ in the approximately categorical form by Gumbel softmax~\cite{jang_iclr2017}.

\mypar{Decoder} 
The functionality of the decoder to predict the future 3D joint positions conditioned on the A-links inferred by the encoder and previous poses; that is,
\begin{equation*}
\mathbf{X}_{t+1}={\rm decode}(\mathbf{X}_t, \dots, \mathbf{X}_{1}, \mathcal{A}) \in \mathbb{R}^{n \times 3},
\end{equation*}
where $\mathbf{X}_{t}$ is the 3D joint positions at the $t$-th frame. The basic idea is to first extract joint features based on the A-links and then convert joint features to future joint positions. Let $\mathbf{x}_{i}^t  \in  \mathbb{R}^d$ be the  features of the $i$th joint at the $t$-th frame. The mapping ${\rm decode} (\cdot)$ works as
\begin{eqnarray*}
(a)~~~\, \mathbf{Q}_{i,j}^{t} 
    &=& \textstyle\sum_{c=1}^C \mathcal{A}_{i,j,c} f_{e}^{(c)}(
    f_{v}^{(c)}(\mathbf{x}_i^t)\oplus f_{v}^{(c)}(\mathbf{x}_j^t))\\
(b)~~~~~ \mathbf{p}_{i}^{t}~\; &=& \textstyle\mathcal{F}(\mathbf{Q}_{i,:} ^{t})\oplus \mathbf{x}_i^t\\
(c)~~~    \mathbf{S}_{i}^{t+1} &=& \text{GRU}(\mathbf{S}_{i}^{t}, \mathbf{p}_{i}^t)\\
(d)~~~    \hat{\mu}_{i}^{t+1} &=& f_{\rm out}(\mathbf{S}_{i}^{t+1}) \in \mathbb{R}^3,
\end{eqnarray*}
where $f_{v}^{(c)}(\cdot)$, $f_{e}^{(c)}(\cdot)$ and $f_{\rm out}(\cdot)$ are MLPs. Step (a) generates link features by weighted averaging on the linking probabilities $\mathcal{A}_{i,j, :} $; Step (b) aggregates the link features to obtain the corresponding joint features; Step (c) uses a gated recurrent unit  (GRU) to update the  joint features~\cite{DBLP:journals/corr/ChoMBB14}; and Step (d) predicts the mean of future  joint positions. We finally sample the future  joint positions $\hat{\mathbf{x}}_{i}^{t+1} \in \mathbb{R}^3$ from a Gaussian distribution, i.e. $\hat{\mathbf{x}}_{i}^{t+1} \sim \mathcal{N}(\hat{\mu}_{i}^{t+1}, \sigma^2 \mathbf{I})$, where $\sigma^2$ denotes the variance and $\mathbf{I}$ is an identity matrix.

We pretrain AIM for a few epoches to warm-up A-links. Mathematically, the cost function of AIM is
\begin{equation*}
    \mathcal{L}_{\rm AIM}(\mathcal{A}) =
    -\sum_{i=1}^n \sum_{t=2}^T \frac{\|\mathbf{x}_i^t-\hat{\mu}_i^t\|^2}{2\sigma^2}
    +\sum_{c=1}^{C}\log\frac{\mathcal{A}_{:,:,c}}{\mathcal{A}^{(0)}_{:,:,c}},
\end{equation*}
where $\mathcal{A}^{(0)}_{:,:,c}$ is the prior of $\mathcal{A}$. In experiments, we find the performance boosts when $p(\mathcal{A})$ promotes the sparsity. The intuition behind is that too many links would capture useless dependencies to confuse action pattern learning; however, in~\eqref{eq:aim_encode_output}, we ensure that $\sum_{c=1}^C \mathcal{A}_{i,j,c} = 1$. Since the probability one is allocated to $C$ link types, it is hard to promote sparsity when $C$ is small. To control the sparsity level, we introduce a ghost link with a large probability, indicating that two joints are not connected through any A-link. The ghost link still ensures that the probabilities sum up to one; that is, for $\forall i,j$, $\mathcal{A}_{i,j, 0} + \sum_{c=1}^C \mathcal{A}_{i,j,c} = 1$, where $\mathcal{A}_{i,j, 0}$ is the probability of isolation. Here we set the prior $\mathcal{A}^{(0)}_{:,:, 0} = P_0$ and $\mathcal{A}^{(0)}_{:,:, c} = P_0/C$ for $c=1, 2, \cdots, C$. In the training of AIM, we only update the probabilities of A-links $\mathcal{A}_{i,j,c}$, where $c=1, \cdots, C$. 

We accumulate $\mathcal{L}_{\rm AIM}$ for multiple samples and minimize it to obtain a warmed-up $\mathcal{A}$. Let $\mathbf{A}_{\rm{act}}^{(c)} = \mathcal{A}_{:,:,c}  \in [0,1]^{n\times n}$  be the $c$-th type of linking probability, which  represents the topology of the $c$-th actional graph. We define the actional graph convolution (AGC), which uses the A-links to capture the actional dependencies among joints. In the AGC, we use $\hat{\mathbf{A}}^{(c)}_{\rm{act}}$ as the graph convolution kernel, where $\hat{\mathbf{A}}_{\rm{act}}^{(c)}= {\mathbf{D}_{\rm{act}}^{(c)}}^{-1}\mathbf{A}_{\rm{act}}^{(c)}$. Given the input $\mathbf{X}_{\rm{in}}$, the AGC is
\begin{eqnarray}
     \label{eq:actional_graph_convolution}
    \mathbf{X}_{\rm{act}}
	& = & {\rm AGC} \left( \mathbf{X}_{\rm{in}} \right)
    \\ \nonumber        
    & = & \sum_{c=1}^{C}\hat{\mathbf{A}}_{\rm{act}}^{(c)}\mathbf{X}_{\rm{in}}\mathbf{W}_{\rm{act}}^{(c)}
     \in  \mathbb{R}^{n\times d_{\rm out}},
\end{eqnarray}
where $\mathbf{W}_{\rm{act}}^{(c)}$ is the trainable weight to capture feature importance. Note that we use the AIM to warm-up A-links in the pretraining process; during the training of action recognition and pose prediction, the A-links are further optimized by forward-passing the encoder of AIM only.

\subsection{Structural Links (S-links)}
As shown in~\eqref{eq:st_gcn_gcn},  $\widetilde{\mathbf{A}^{(p)}}  \mathbf{X}_{\rm{in}}$ aggregates the $1$-hop neighbors' information in skeleton graph; that is, each layer in ST-GCN only diffuse information in a local range. To obtain long-range links, we use the high-order polynomial of ${\bf A}$, indicating the S-links. Here we use $\hat{\mathbf{A}}^{L}$ as the graph convolution kernel, where $\hat{\mathbf{A}} =\mathbf{D}^{-1}\mathbf{A}$ is the graph transition matrix and $L$ is the polynomial order. $\hat{\mathbf{A}}$ introduces the degree normalization to avoid the magnitude explosion and has probabilistic intuition~\cite{ilprints361, DBLP:journals/tsp/ChenTFVK18}. With the $L$-order polynomial, we define the structural graph convolution (SGC), which can directly reach the $L$-hop neighbors to increase the receptive field. The SGC is formulated as
\begin{eqnarray}
\label{eq:structural_graph_convolution}
    \mathbf{X}_{\rm{struc}}
    & = & {\rm SGC} \left( \mathbf{X}_{\rm{in}} \right)
    \\ \nonumber
     & = & \sum_{l=1}^{L} \sum_{p \in \Pset{}}
      \mathbf{M}_{\rm{struc}}^{(p, l)}
         \circ  \hat{\mathbf{A}}^{(p)l}
       \mathbf{X}_{\rm{in}}
    \mathbf{W}_{\rm{struc}}^{(p, l)} 
    \\ \nonumber
    & \in & \mathbb{R}^{n \times d_{\rm out}},
\end{eqnarray}
where $l$ is the polynomial order, $\hat{\mathbf{A}}^{(p)}$ is the graph transition matrix for $p$-th parted graph, $\mathbf{M}_{\rm{struc}}^{(p, l)} \in \mathbb{R}^{n \times n}$ and $\mathbf{W}_{\rm{struc}}^{(p, l)} \in \mathbb{R}^{n \times d_{\rm struc}}$ are the trainable weights to capture edge weights and feature importance; namely, larger weight indicates more important corresponding feature. The weights are introduced for each polynomial order and each individual parted graph. Note that with the degree normalization, the graph transition matrix $\hat{\mathbf{A}}^{(p)}$ provides the nice initialization for edge weights, which stabilizes the learning of $\mathbf{M}_{\rm{struc}}^{(p, l)}$. When $L=1$, the SGC degenerates to the original spatial graph convolution operation. For $L>1$, the SGC acts like the Chebyshev filter and is able to approximate the convolution designed in the graph spectral domain~\cite{DBLP:journals/corr/BronsteinBLSV16}


\subsection{Actional-Structural Graph Convolution Block}
To integrally capture the actional and structural features among arbitrary joints, we combine the AGC and SGC and develop the actional-structural graph convolution (ASGC). In~\eqref{eq:actional_graph_convolution} and~\eqref{eq:structural_graph_convolution}, we obtain the joint features from AGC and SGC in each time stamp, respectively. We use a convex combination of both as the response of the ASGC. Mathematically, the ASGC operation is formulated as
\begin{eqnarray*}
  \mathbf{X}_{\rm{out}} 
& = & {\rm ASGC} \left( \mathbf{X}_{\rm{in}} \right)
  \\ \nonumber
  & = &
   \mathbf{X}_{\rm{struc}} + \lambda \mathbf{X}_{\rm{act}}  \in \mathbb{R}^{n \times d_{\rm out}},
\end{eqnarray*}
where $\lambda$ is a hyper-parameter, which trades off the importance between structural features and actional features. A non-linear activation function, such as $\rm{ReLU(\cdot)}$, can be further introduced after ASGC.

\noindent{\textbf{Theorem 1.}
The actional-structural graph convolution is a valid linear operation; that is, when $\mathbf{Y}_{1}  =  {\rm ASGC} \left(\mathbf{X}_{1} \right)$ and $ \mathbf{Y}_{2}  =  {\rm ASGC} \left(\mathbf{X}_{2} \right)$. Then, 
$ a  \mathbf{Y}_{1}  + b  \mathbf{Y}_{2} 
 =  {\rm ASGC} \left( a \mathbf{X}_{1} + b  \mathbf{X}_{2}   \right)$,   $\forall a, b \in \mathbb{R}$.}

The linearity ensures that ASGC effectively preserves information from both structural and actional aspects; for example, when the response from the action aspect is stronger, it can be effectively reflected through ASGC.

\begin{figure}
\centering
\includegraphics[width=7.5cm]{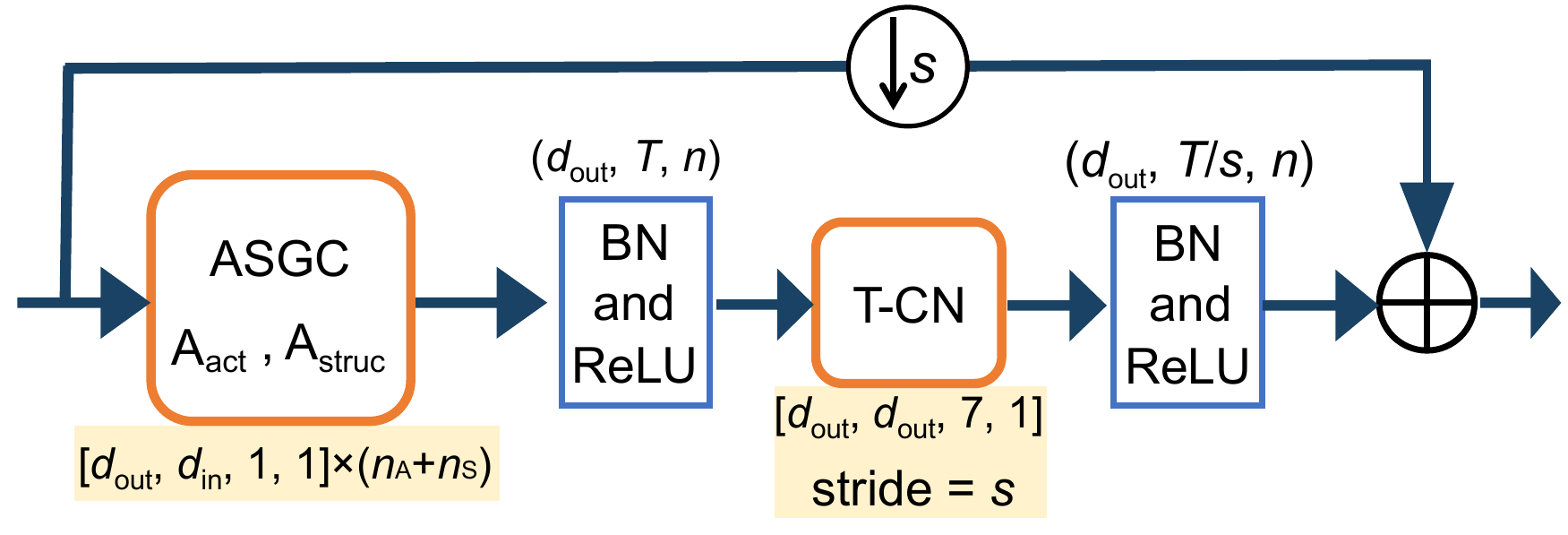} 
\caption{\small An AS-GCN block consists of ASGC, T-CN, and other operations: batch normalization (BN), ReLU and the residual block. The shapes of data are above the BN and ReLU blocks. The shapes of network parameters are under ASGC and T-CN.}
\label{fig:AS_GCN_block}
\end{figure}

To capture the inter-frame action features, we use one layer of temporal convolution (T-CN) along the time axis, which extracts the temporal feature of each joint independently but shares the weights on each joint.  Since ASGC and T-CN learns spatial and temporal features, respectively, we concatenate both layers as an actional-structural graph convolution block (AS-GCN block) to extract temporal features from various actions; see Figure~\ref{fig:AS_GCN_block}. Note that ASGC is a single operation to extract only spatial information and the AS-GCN block includes a series of operations to extract both spatial and temporal information.

\subsection{Multitasking of AS-GCN}
\begin{figure}
\centering
\includegraphics[width=7.5cm]{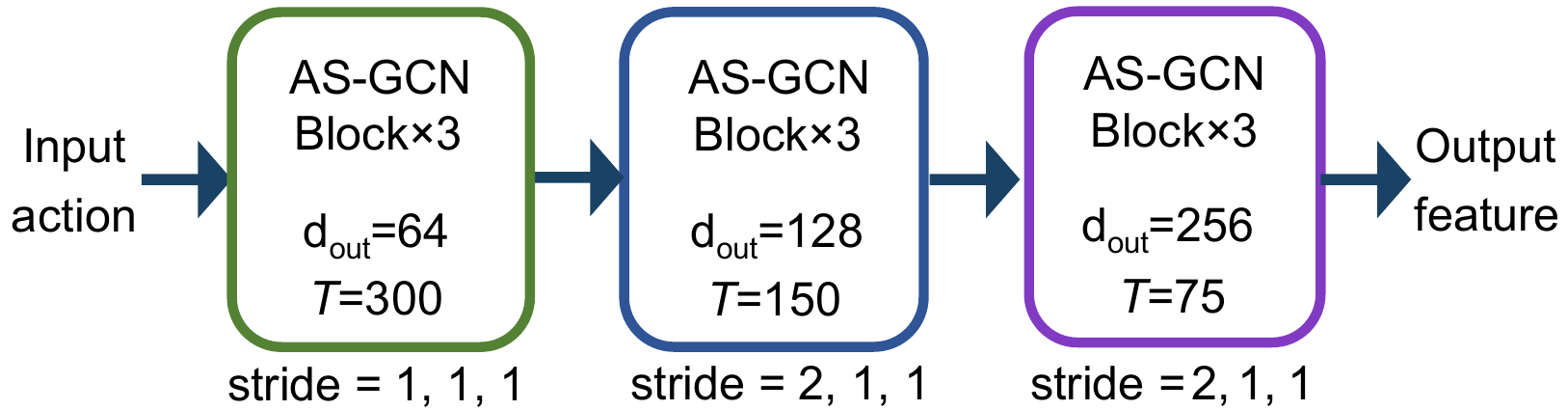} 
\caption{\small The backbone network of AS-GCN, which includes nine AS-GCN blocks. The feature dimensions are presented.}
\label{fig:backbone}
\end{figure}

\textbf{Backbone network.}
We stack a series of AS-GCN blocks to be the backbone network, called  AS-GCN; see Figure~\ref{fig:backbone}. After the multiple spatial and temporal feature aggregations, AS-GCN extracts high-level semantic information across time.

\textbf{Action recognition head.}
To classify actions, we construct a recognition head following the backbone network. We apply the global averaging pooling on the joint and temporal dimensions of the feature maps output by the backbone network, and obtain the feature vector, which is finally fed into a softmax classifier to obtain the predicted class-label  $\hat{\mathbf{y}}$. The loss function for action recognition is the standard cross entropy loss
\begin{equation*}
    \mathcal{L}_{\rm{recog}} = -\mathbf{y}^{\rm{T}}\log(\hat{\mathbf{y}}),
\end{equation*}
where $\mathbf{y}$ is the ground-truth label of the action.

\textbf{Future pose prediction head.} 
Most previous works on the analysis of skeleton data focused on the classification task. Here we also consider pose prediction; that is, using AS-GCN to predict future 3D joint positions given by historical skeleton-based actions.

\begin{figure}
\centering
\includegraphics[width=8.3cm]{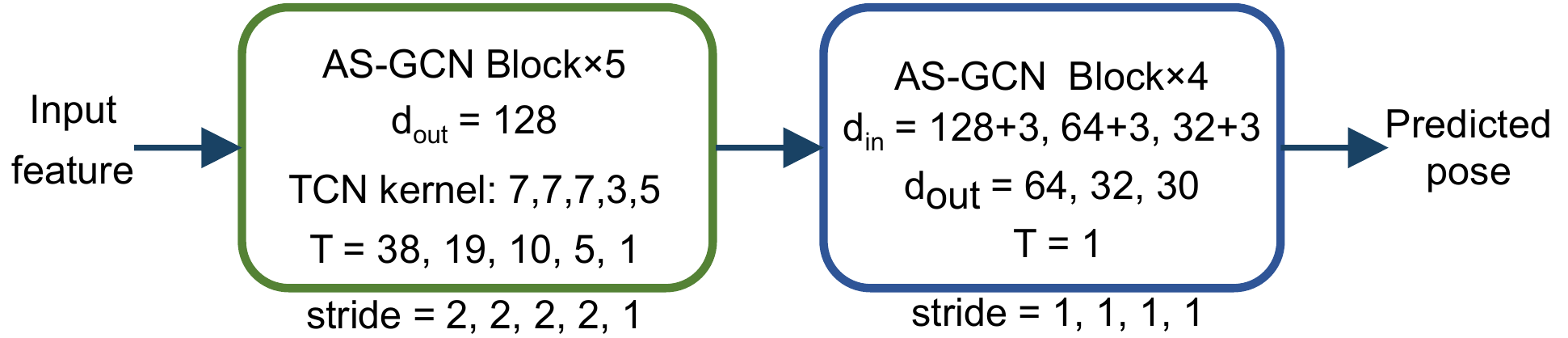} 
\caption{\small Future prediction head of AS-GCN.}
\label{fig:predhead}
\end{figure}

To predict future poses, we construct a prediction module followed by the backbone network. We use several AS-GCN blocks to decode the high-level feature maps extracted from the historical data and obtain the predicted future 3D joint positions
$\hat{\mathcal{X}} \in \mathbb{R}^{n \times 3 \times T'}$; see Figure~\ref{fig:predhead}.
The loss function for future prediction is the standard $\ell_2$ loss
\begin{equation}
    \mathcal{L}_{\rm{predict}} = \frac{1}{n  d T'} \sum_{i=1}^{n d}\sum_{t = 1}^{T'} \left\| \hat{\mathcal{X}}_{i, :, t}-\mathcal{X}_{i, :, t} \right\|_2^2.
\end{equation}

\mypar{Joint model}
In practice, when we train the recognition head and future prediction head together, recognition performance gets improved. The intuition behind is that the future prediction module promotes self-supervision and avoids overfitting in recognition.

\section{Experiments}

\subsection{Data sets and Model Configuration}
\textbf{NTU-RGB+D.} NTU-RGB+D, containing $56,880$ skeleton action sequences completed by one or two performers and categorized into $60$ classes, is one of the largest data sets for skeleton-based action recognition \cite{Shahroudy_2016_CVPR}. It provides the 3D spatial coordinates of 25 joints for each human in an action. For evaluating the models, two  protocols are recommended: Cross-Subject and Cross-View. In Cross-Subject, $40,320$ samples performed by $20$ subjects are separated into training set, and the rest belong to test set. Cross-View assigns data according to camera views, where training and test set have $37,920$ and $18,960$ samples, respectively.

\textbf{Kinetics.} Kinetics is a large data set for human action analysis, containing over $240,000$ video clips \cite{DBLP:journals/corr/KayCSZHVVGBNSZ17}. There are 400 types of actions. Due to only RGB videos are provided, we obtain skeleton data by estimating joint locations on certain pixels with OpenPose toolbox~\cite{Cao_2017_CVPR}. The toolbox generates 2D pixel coordinates $(x,y)$ and confidence score $c$ for totally $18$ joints from the resized videos with resolution of $340\times256$. We represent each joint as a three-element feature vector: $[x,y,c]^{\mathrm{T}}$. For the multiple-person cases, we select the body with the highest average joint confidence in each sequence. Therefore, one clip with $T$ frames is transformed into a skeleton sequence with dimension of $18\times3\times T)$. Finally, we pad each sequence by repeating the data from the start to totally $T=300$.

\textbf{Model Setting.} We construct the backbone of AS-GCN with 9 AS-GCN blocks, where the features dimensions are 64, 128, 256 in each three blocks. The structure and operations of future pose prediction module are symmetric to the recognition module and we use the residual connection. In the AIM, we set the hidden features dimensions to be 128. The number of A-link types $C=3$ and the prior of the ghost link $P_0=0.95$. $\lambda=0.5$. We use PyTorch 0.4.1 and train the model for 100 epochs on 8 GTX-1080Ti GPUs. The batch size is 32. We use the SGD algorithm to train both recognition and prediction heads of AS-GCN, whose learning rate is initially 0.1, decaying by 0.1 every 20 epochs. We use Adam optimizer \cite{Kingma_iclr2015} to train the A-link inference module with the initial learning rate $0.0005$. All hyper-parameters are selected using a validation set.

\subsection{Ablation Study}
To analyze each individual component of AS-GCN, we conduct extensive experiments on Cross-Subject benchmark of the NTU-RGB+D data set~\cite{Shahroudy_2016_CVPR}.

\textbf{Effect of link types.} Here we focus on validating the proposed A-links and S-links.
In the experiments, we consider three link-type combinations, including S-links, A-links and AS-links (A-links $+$ S-links), with the original skeleton links. While involving S-links, we respectively set the polynomial order $L = 1,2,3,4$ in the model. Note that when $L = 1$, the corresponding S-link is exactly the skeleton itself.
\begin{table}[]
    \small
    \centering
    \caption{\small The recognition accuracy on NTU-RGB+D Cross-Subject with various links: S-links, A-links and A- with S-links (AS-links). We tune the polynomial order in S-links from 1 to 4.}
    \begin{tabular}{c|c|c|c}
       \hline
       Polynomial order & S-links & A-links &  AS-links \\
       \hline
       1 & 81.5\% & \multirow{4}{*}{83.2\%}  & 83.2\%  \\
       2 & 82.2\% &  & 83.7\%\\
       3 & 83.4\% &  & 84.4\%\\
       4 & 84.2\% &  & \textbf{86.1\%}\\
       \hline
    \end{tabular}
    \label{tab:links}
    \vspace{-10pt}
\end{table}
Table~\ref{tab:links} shows the classification accuracy of action recognition. We see that (1) either S-links with higher polynomial order or A-links can improve the recognition performance; (2) when using both S-links and A-links together, we achieve the best performance; (3) with only A-links and skeleton graphs, the classification accuracy result reaches $83.2\%$, which is higher than S-links with polynomial order 1 ($81.5\%$). These results validate the limitation of the original skeleton graph and the effectiveness of the proposed S-link and A-link.

\begin{figure}
\centering
\includegraphics[width=7.5cm]{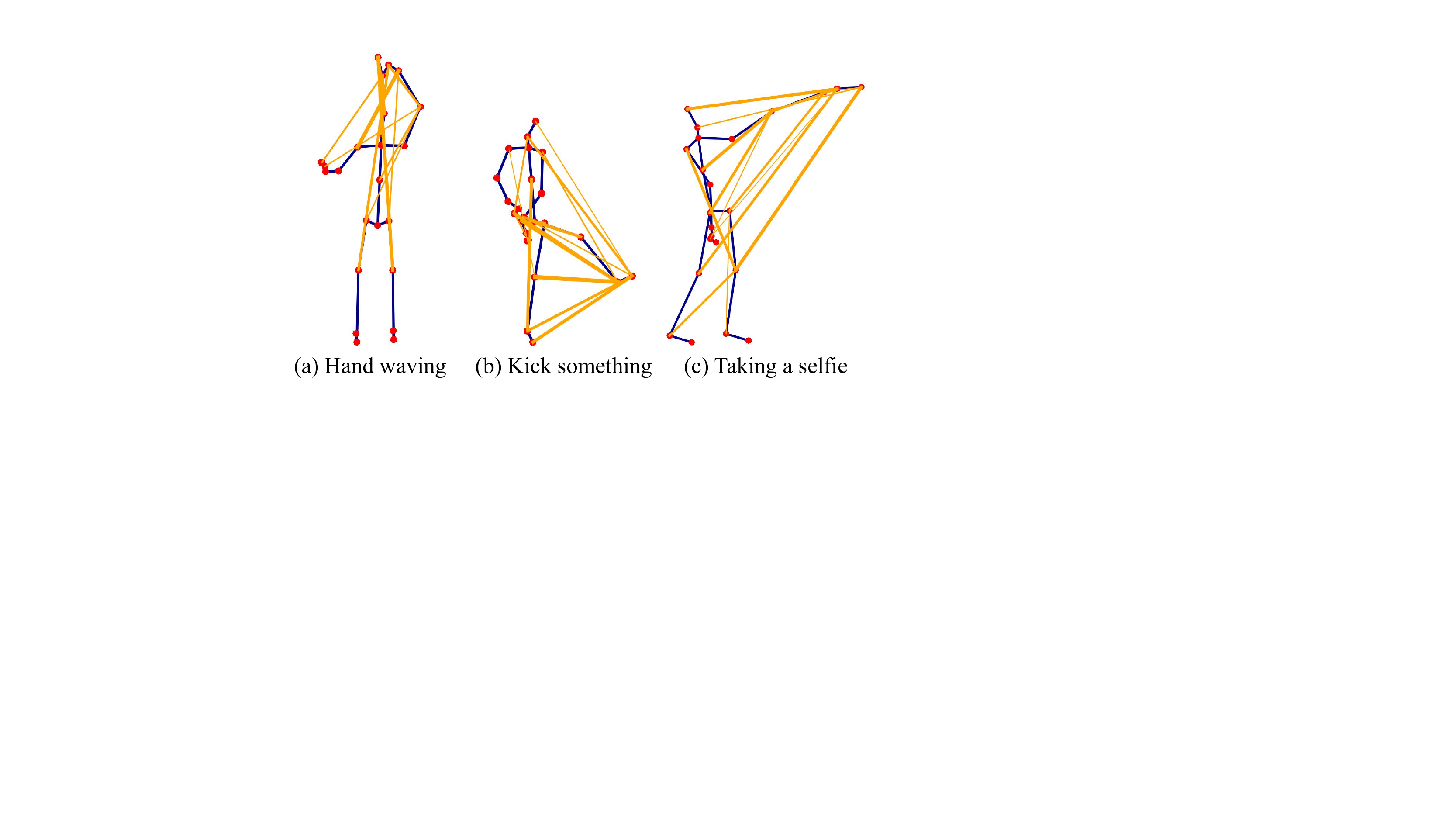} 
\caption{\small A-links in actions. We plot the A-links with probabilities larger than 0.9. The wider line denotes the larger probability.}
\label{fig:3} 
\end{figure}

\textbf{Visualizing A-links.} Various actions may activate different actional dependencies among joints. Figure~\ref{fig:3} shows the inferred A-links for three actions. The A-links with probabilities larger than 0.9 are illustrated as orange lines, where wider lines represent larger linking probabilities. 
We see that (1) in Plots (a) and (c), the actions of hand waving and taking a selfie are mainly upper limb actions, where arms have large movements and interact with the whole bodies, so that many A-links are built between arms and other body parts. (2) In Plot (b), the action of kicking something shows that the kicked leg is highly correlated to the other joints, indicating the body balancing during this action. These results validate that richer information of action patterns is captured by A-links.

\textbf{The number and priors of A-links. } To select the appropriate $C$: the number of A-link types; and $P_0$: the prior of the ghost links for training the AIM. We test the models with different $C$ and $P_0$ to obtain the corresponding recognition accuracies, which are presented in Table~\ref{tab:acc-hyp}.
\begin{table}[tb]
    \centering
    \small
    \caption{\small Recognition accuracy with various number of A-link types and different priors of the ghost links.}
    \begin{tabular}{c|ccccc}
      \hline
        $C$ & 1 & 2 & 3 & 4 & 5\\
      \hline
        Acc & 84.6\% & 86.5\% & {\bf 86.8\%} & 85.8\% & 83.3\%\\
      \hline
        $P_0$ &0.99 & 0.95 & 0.50 & 0.20 & 0.00\\
      \hline
        Acc &86.0\%& {\bf 86.8\%} & 84.3\% & 82.7\% & 81.1\%\\
      \hline
    \end{tabular}
    \label{tab:acc-hyp}
\end{table}
We see that when $C=3$ and $P_0=0.95$, we could obtain the highest recognition accuracy. The intuition is that too few A-link types cannot capture significant actional relationships and too many causes overfitting. And the sparse A-links would promote the recognition performance.

\textbf{Effect of prediction head.} To analyze the effect of the prediction head on improving recognition performance, we perform two groups of contrast tests. For the first group, AS-GCNs only employs S-links for action recognition but one has prediction head and the other does not have. In the other group, AS-GCN with/without prediction head additionally employ A-links. The polynomial order of S-links is from 1 to 4. 
\begin{table}[]
\small
    \centering
    \caption{\small The recognition results of models with/without prediction heads on NTU-RGB+D Cross-Subject are listed, where the models use AS-links. We tune the order of S-links from 1 to 4.}
    \begin{tabular}{c|cc}
       \hline
       Polynomial order & AS-links & + Pred\\
       \hline
       1  &  83.2\% & 84.0\%\\
       2  &  83.7\% & 84.3\%\\
       3  &  84.4\% & 85.1\%\\
       4  &  86.1\% & \textbf{86.8\%}\\
       \hline
    \end{tabular}
    \label{tab:recons}
    \vspace{-10pt}
\end{table}
Table~\ref{tab:recons} shows the classification results with/without prediction heads. We obtain better recognition performance consistently by around $1\%$ when we introduce the prediction head. The intuition behind is that the prediction modules promotes to preserve more detailed information and introduce self-supervision to help recognition module avoid overfitting and achieve higher action recognition performance. The sparse skeleton actions may sometimes rely on the detailed motions rather than coarse profiles which are easily-confused in some actions classes.

\begin{figure} 
\centering
\includegraphics[width=7.2cm]{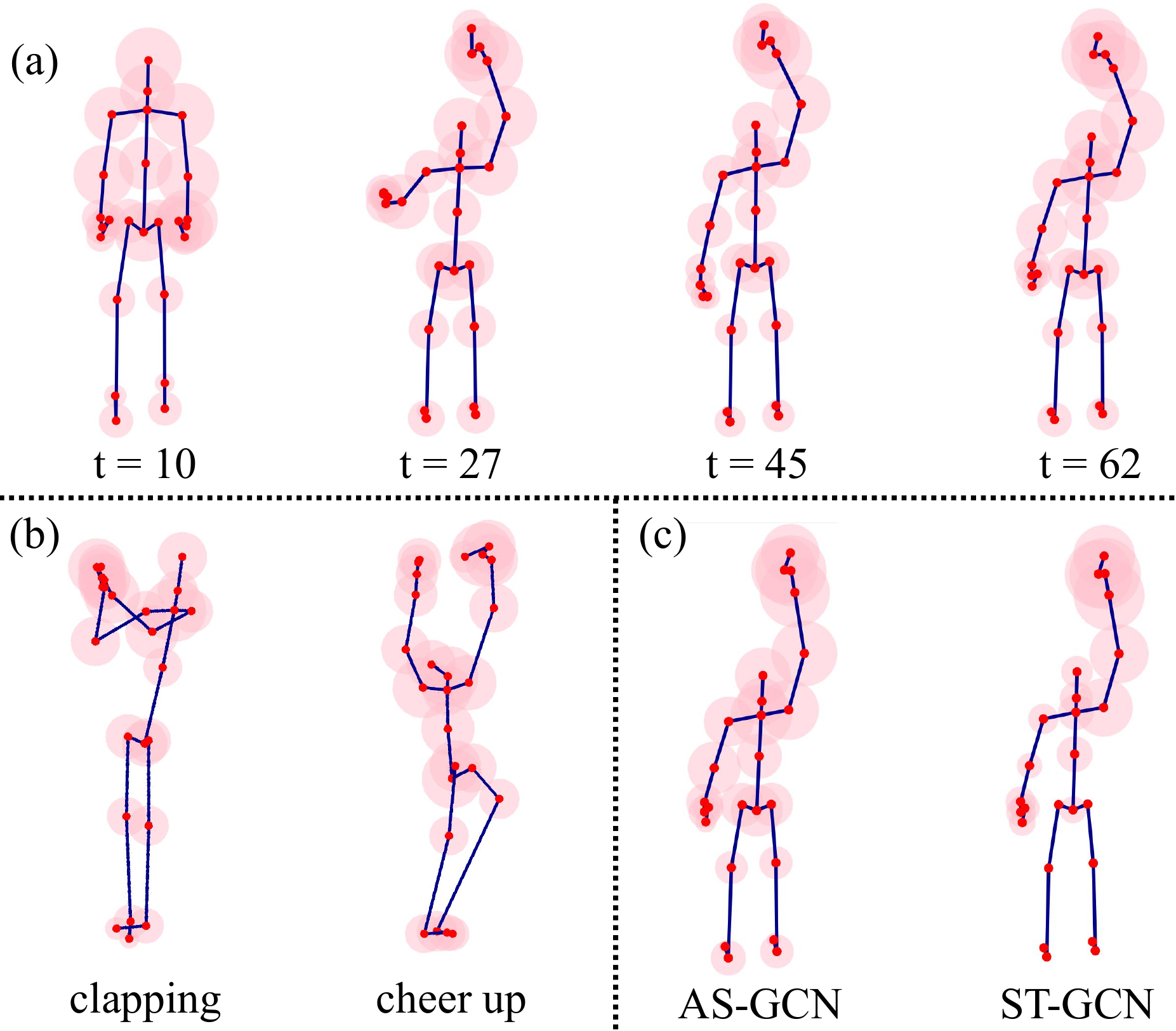} 
\caption{\small Feature responses in the last layer of AS-GCN backbone. The areas of translucid circles indicates the response magnitudes. Plot (a) shows the feature maps of action 'hand waving' in different frames; Plot (b) shows the feature maps of other two actions; Plot (c) compares AS-GCN to ST-GCN on 'hand waving'.}
\label{fig:feat_map} 
\end{figure}

\textbf{Feature visualization.} To validate how the features of each joint effect on the final performance, we visualize the feature maps of actions in Figure~\ref{fig:feat_map}, where the circle around each joint indicates magnitude of feature responses of this joint in the last AS-GCN block of the recognition module of AS-GCN.
Plot (a) shows the feature responses of the action 'hand waving' at different time. At the initial phase of action, namely Frame 15, many joints of upper limb and trunk have approximately comparative responses; however, in the subsequent frames, large responses are distributed on the upper body especially waving arm. Note that other non-functional joints are not much neglected, because abundant hidden relationships are built. Plot (b) shows the other two actions, where we are able to capture many long-range dependencies. Plot (c) compares the features between AS-GCN and ST-GCN. ST-GCN does apply multi-layer GCNs to cover the entire spatial domain; however, the feature are weakened during the propagation and distant joints cannot interact effectively, leading to localized feature responses. On the other hand, the proposed AS-GCN could capture useful long-range dependencies to recognize the actions.

\subsection{Comparisons to the State-of-the-Art}
We compare AS-GCN on skeleton-based action recognition tasks  with the state-of-the-art methods on the data sets of NTU-RGB+D and Kinetics. On NTU-RGB+D, we train AS-GCN on two recommended benchmarks: Cross-Subject and Cross-View, then we respectively obtain the top-1 classification accuracies in the test phase. We compare with covering hand-crafted methods \cite{Vemulapalli_2014_CVPR}, RNN/CNN-based deep learning models \cite{Du_2015_CVPR, Shahroudy_2016_CVPR, 10.1007/978-3-319-46487-9_50, a8014941, vis_cnn, Ke_2017_CVPR, ijcai_ChaoLi} and recent graph-based methods \cite{AAAI1817135, Tang_2018_CVPR, Si_2018_ECCV}. Specifically, ST-GCN \cite{AAAI1817135} combines GCN with temporal CNN to capture spatio-temporal features, and SR-TSL \cite{Si_2018_ECCV} use gated recurrent unit (GRU) to  propagate messages on graphs and use LSTM to learn the temporal features. Table \ref{tab:ntu} shows the comparison. We see that the proposed AS-GCN outperforms the other methods.
\begin{table}[]
\small
\centering
\caption{\small Comparison of action recognition performance on NTU-RGB+D. The classification accuracies on both Cross-Subject and Cross-View benchmarks are presented.}
\begin{tabular}{c|cc}
    \hline
     Methods & Cross Subject & Cross View \\
    \hline
     Lie Group \cite{Vemulapalli_2014_CVPR} & 50.1\% & 52.8\% \\
     H-RNN \cite{Du_2015_CVPR} & 59.1\% & 64.0\% \\
     Deep LSTM \cite{Shahroudy_2016_CVPR} & 60.7\% & 67.3\% \\
     PA-LSTM \cite{Shahroudy_2016_CVPR} & 62.9\% & 70.3\% \\
     ST-LSTM+TS \cite{10.1007/978-3-319-46487-9_50} & 69.2\% & 77.7\% \\
     Temporal Conv \cite{a8014941} & 74.3\% & 83.1\% \\
     Visualize CNN \cite{vis_cnn} & 76.0\% & 82.6\% \\
     C-CNN+MTLN \cite{Ke_2017_CVPR} & 79.6\% & 84.8\% \\
     ST-GCN \cite{AAAI1817135} & 81.5\% & 88.3\% \\
     DPRL~\cite{Tang_2018_CVPR} & 83.5\% & 89.8\% \\
     SR-TSL \cite{Si_2018_ECCV} & 84.8\% & 92.4\% \\
     HCN~\cite{ijcai_ChaoLi} & 86.5\% & 91.1\% \\
    \hline
     AS-GCN (Ours) & \textbf{86.8\%} & \textbf{94.2\%} \\
    \hline
\end{tabular}
\label{tab:ntu}
\end{table}

\begin{table}[]
\small
\centering
\caption{\small Comparison of action recognition performance on Kinetics. We list the top-1 and top-5 classification accuracies.}
\begin{tabular}{c|cc}
    \hline
     Methods & Top-1 Acc & Top-5 Acc \\
    \hline
     Feature Enc \cite{Fernando_2015_CVPR} & 14.9\% & 25.8\%\\
     Deep LSTM \cite{Shahroudy_2016_CVPR} & 16.4\% & 35.3\%\\
     Temporal Conv \cite{a8014941} & 20.3\% & 40.0\%\\
     ST-GCN \cite{AAAI1817135} & 30.7\% & 52.8\%\\
    \hline
     AS-GCN (Ours) & \textbf{34.8\%} & \textbf{56.5\%} \\
    \hline
\end{tabular}
\label{tab:kinetics}
\vspace{-10pt}
\end{table}
In the Kinetics dataset, we compare AS-GCN with four state-of-the-art approaches. A hand-crafted based method named Feature Encoding \cite{Fernando_2015_CVPR} is presented at first. Then Deep LTSM and Temporal ConvNet \cite{Shahroudy_2016_CVPR, a8014941} are implemented as two deep learning models on Kinetics skeletons. Additionally, ST-GCN is also evaluated for Kinetics action recognition. Table \ref{tab:kinetics}  shows the top-1 and top-5 classification performances.
We see that AS-GCN outperforms the other competitive methods in both top-1 and top-5 accuracies.

\subsection{Future Pose Prediction}
We evaluate the performance of  AS-GCN for future pose prediction. For each action, we take all frames except for the last ten as the input. We attempt to predict the last ten frames. 
\begin{figure}
\centering
\includegraphics[width=7.cm]{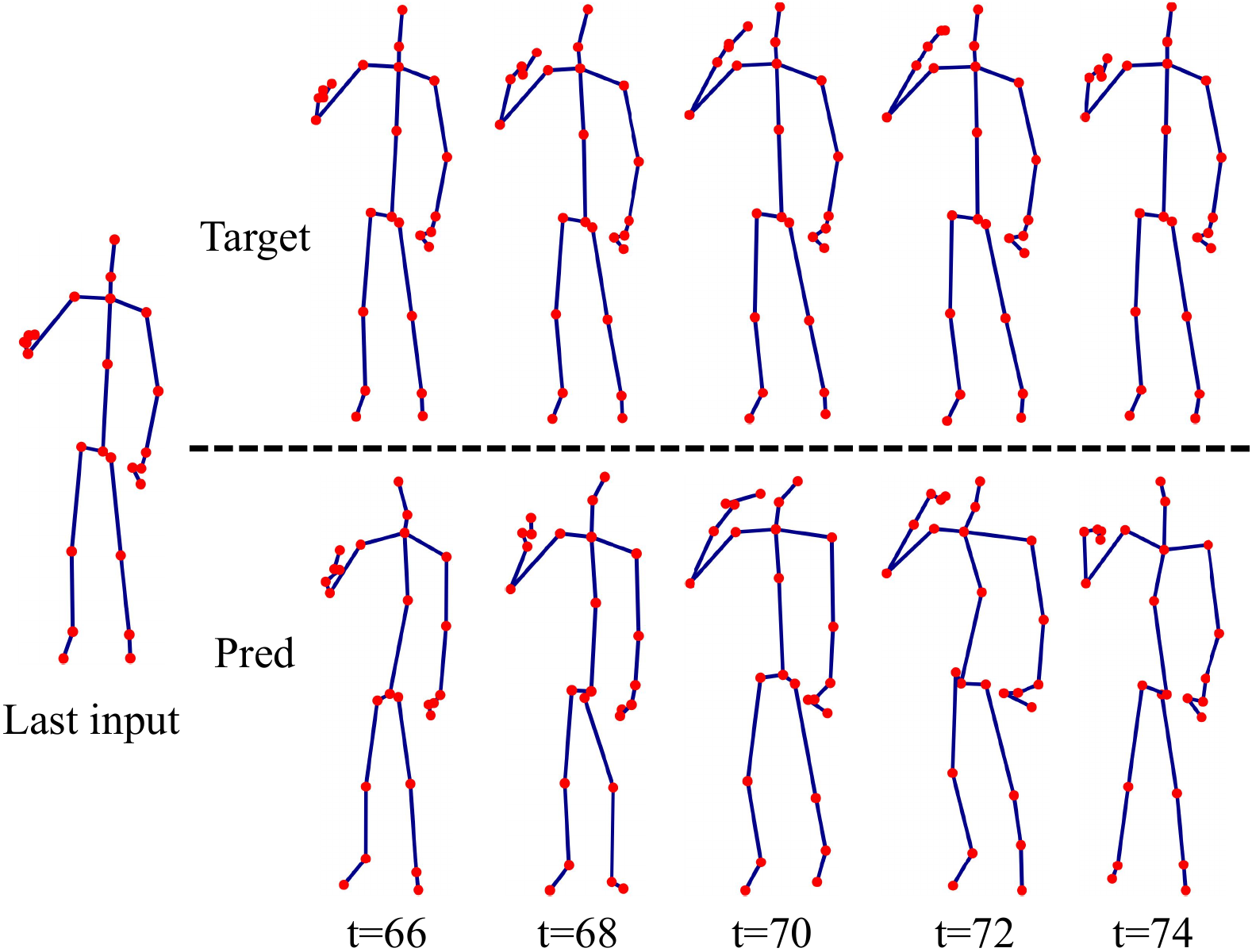} 
\caption{\small The predicted action samples from prediction module. We present the action ``Use a fan'' in NTU-RGB+D dataset. Both ground-truth and predicted data are shown.}
\label{fig:action_recon} 
\end{figure}
Figure~\ref{fig:action_recon} visualizes the original and predicted action. We sample five frames at regular intervals in ten. The predicted frame provides the future joint position with a low error, especially the characteristic actional body parts, e.g. shoulders and arms. As for the peripheral parts such as legs and feet, the predicted positions have larger error, which is the secondary information of the action pattern. These results show that AS-GCN preserves more detailed features especially for the action-functional joints.

\section{Conclusions}
We propose the actional-structural graph convolution networks (AS-GCN) for skeleton-based action recognition. The A-link inference module captures actional dependencies. We also extend the skeleton graphs to represent higher order relationships. The generalized graphs are fed to AS-GCN block for a better representation of actions. An additional future pose prediction head captures more detailed patterns through self-supervision. We validate AS-GCN in action recognition using two data sets, NTU-RGB+D and Kinetics. The AS-GCN achieves large improvement compared with the previous methods. Moreover, AS-GCN also shows promising results for future pose prediction.

\section*{Acknowledgement}
We are supported by The High Technology Research and Development Program of China (2015AA015801), NSFC (61521062), and STCSM (18DZ2270700).

\newpage
{\small
\bibliographystyle{ieee}
\bibliography{egbib}
}

\newpage
\section*{Appendix A: Theorem Proof}

\begin{myThm}  
The actional-structural graph convolution is a valid linear operation; that is, when $\mathbf{Y}_{1}  =  {\rm ASGC} \left( \mathbf{X}_{1} \right)$ and $ \mathbf{Y}_{2}  =  {\rm ASGC} \left( \mathbf{X}_{2} \right)$. Then, 
$ a  \mathbf{Y}_{1}  + b  \mathbf{Y}_{2} 
 =  {\rm ASGC} \left( a \mathbf{X}_{1} + b  \mathbf{X}_{2}   \right)$, $\forall a, b$. 
\end{myThm}

\begin{prf}
The operations in actional graph convolution (AGC) are all linear, as well as the structural graph convolution (SGC). The AGC satisfies
\begin{eqnarray*}
& &{\rm AGC}(a\mathbf{X}_1+b\mathbf{X}_2) \\
\nonumber
&=& \sum_{c=1}^{C}\hat{\mathbf{A}}_{\rm{act}}^{(c)}\left(a\mathbf{X}_1+b\mathbf{X}_2\right)\mathbf{W}_{\rm{act}}^{(c)}\\
\nonumber
&=& a \left(\sum_{c=1}^{C}\hat{\mathbf{A}}_{\rm{act}}^{(c)}\mathbf{X}_1\mathbf{W}_{\rm{act}}^{(c)}\right)+b\left(\sum_{c=1}^{C}\hat{\mathbf{A}}_{\rm{act}}^{(c)}\mathbf{X}_2\mathbf{W}_{\rm{act}}^{(c)}\right)\\
&=& a{\rm AGC}(\mathbf{X}_1)+b{\rm AGC}(\mathbf{X}_2).
\end{eqnarray*}
Similarly, SGC satisfies
\begin{eqnarray*}
&& {\rm SGC}(a\mathbf{X}_1+b\mathbf{X}_2)\\
\nonumber
&=& \sum_{\ell=1}^{L}\sum_{p\in\mathcal{P}}\mathbf{M}_{\rm struct}^{(p,\ell)}\circ\hat{\mathbf{A}}^{(p)\ell}(a\mathbf{X}_1+b\mathbf{X}_2)\mathbf{W}_{\rm{struc}}^{(p,\ell)}\\
\nonumber
&=& a{\rm SGC}(\mathbf{X}_1)+b{\rm SGC}(\mathbf{X}_2).
\end{eqnarray*}
With both AGC and SGC operations, the actional-structural convolution (ASGC) is formulated as
\begin{eqnarray*}
  \mathbf{Y}_1 
& = & {\rm ASGC} \left( \mathbf{X}_1 \right)
  \\ \nonumber
  & = &
   (1 - \lambda){\rm SGC} \left( \mathbf{X}_1 \right) + \lambda {\rm AGC} \left( \mathbf{X}_1 \right),
\end{eqnarray*}
which is a linear summation of AGC and SGC. Therefore, we have
\begin{eqnarray*}
& & {\rm ASGC} \left( a\mathbf{X}_1+b\mathbf{X}_2 \right)
  \\ \nonumber
  & = &
   (1 - \lambda){\rm SGC} \left( a\mathbf{X}_1+b\mathbf{X}_2 \right) + \lambda {\rm AGC} \left( a\mathbf{X}_1+b\mathbf{X}_2 \right),
  \\ \nonumber
  & = &
  (1 - \lambda)(a{\rm SGC} \left(\mathbf{X}_1\right)+b{\rm SGC} \left(\mathbf{X}_2 \right)) 
  \\ \nonumber
  & & + \lambda (a {\rm AGC} \left( \mathbf{X}_1\right)+ b {\rm AGC} \left(\mathbf{X}_2 \right)),
  \\ \nonumber
  & = & a((1-\lambda){\rm SGC}(\mathbf{X}_1)+\lambda{\rm AGC}(\mathbf{X}_1))
  \\ \nonumber
  & & + b((1-\lambda){\rm SGC}(\mathbf{X}_2)+\lambda{\rm AGC}(\mathbf{X}_2))
  \\ \nonumber
  & = & a{\rm ASGC}(\mathbf{X}_1)+b{\rm ASGC}(\mathbf{X}_2)
  \\ \nonumber
  & = & a\mathbf{Y}_1+b\mathbf{Y}_2.
\end{eqnarray*}
The ASGC is a linear operation for the input data.

\end{prf}

\section*{Appendix B: Model Architectures}
In this section, we show the detailed architectures of the  proposed AS-GCN model.
\subsection*{A-links Inference Module (AIM)}
\subsubsection*{Encoder}
Given the 3D joint positions of $n$ joints across $T$ frames, $\mathcal{X}\in \mathbb{R}^{n\times3\times T}$, we first downsample the videos to obtain 50 frames from the valid frames at regular intervals. If $T<50$, we pad the sequences to be $50$ frames with $0$. For any joint $v_i$ on the body, where $i\in\{1,2,\dots,n\}$, we represent the joint feature across $50$ frames as $\mathbf{x}_i\in\mathbb{R}^{150}$. We set that there are four types of A-links for actional dependencies (including the ghost link). As for link feature aggregation, we use average operation for all links surrounding one joint. The operations in the encoder in AIM are presented in Table~\ref{tab:AIM_encoder}.
\begin{table}[h]
\centering
\caption{\small The architecture of the encoder in AIM}
\begin{tabular}{|c|c|c|}
\hline
Input & Operation & Output \\
\hline\hline
$\mathbf{p}_i^{(0)}=\mathbf{x}_i$ & $150 \xrightarrow{\text{elu}} 128 \xrightarrow{\text{elu}} 128$ (\text{bn}) & $\mathbf{p}_i^{(1)}$\\\hline

$\mathbf{p}_i^{(1)}$, $\mathbf{p}_j^{(1)}$ & $\mathbf{p}_i^{(1)} \oplus \mathbf{p}_j^{(1)}$ & $\mathbf{Q}_{i,j}^{(2)}$\\\hline

\multirow{2}{*}{$\mathbf{Q}_{i,:}^{(2)}$} & $(\frac{1}{n}\sum_{j=1}^{n}\mathbf{Q}_{i,j}^{(2)})\oplus \mathbf{p}_i^{(1)}$ & \multirow{2}{*}{$\mathbf{p}_i^{(2)}$}\\

 & $ 384 \xrightarrow{\text{elu}} 128 \xrightarrow{\text{elu}} 128$ (\text{bn}) &  \\\hline

$\mathbf{p}_i^{(2)}$, $\mathbf{p}_j^{(2)}$ & $\mathbf{p}_i^{(2)} \oplus \mathbf{p}_j^{(2)}$ & $\mathbf{Q}_{i,j}^{(3)}$\\\hline

$\mathbf{Q}_{i,j}^{(3)}$ & $256 \xrightarrow{\text{elu}} 128 \xrightarrow{\text{elu}} 128$ (\text{bn}) $\to 4$ & $\mathcal{A}_{i,j,:}$\\\hline
\end{tabular}
\label{tab:AIM_encoder}
\end{table}
The activation functions of MLPs in the encoder are exponential linear unit (elu) functions, and 'bn' denotes the batch normalization to the features. $\oplus$ is the concatenation operation.

\subsubsection*{Decoder}
We present the detailed configuration of the decoder of AIM. Given the position of joint $v_i$ at time $t$, $\mathbf{x}_i^{t}$, the decoder aims to predict the future joint position $\mathbf{x}_i^{t+1}$ conditioned on the sourrounding A-links, $\mathcal{A}_{i,j,:}$. The architectures are presented in Table~\ref{tab:AIM_decoder}.
\begin{table}[h]
\centering
\caption{\small The architecture of the decoder in AIM}
\begin{tabular}{|c|c|c|}
\hline
Input & Operation & Output \\
\hline\hline
\multirow{3}{*}{$\mathbf{x}_i^t$, $\mathbf{x}_j^t$} & $3 \xrightarrow{\text{relu}} 64$ $(c)$  & \multirow{3}{*}{$\mathbf{Q}_{i,j}^t$}\\

 & $ \oplus $ & \\

 & $\sum_{c=1}^{C}\mathcal{A}_{i,j,c} \cdot (128 \xrightarrow{\text{relu}} 64$ $(c))$ & \\\hline
 
$\mathbf{Q}_{i,j}^t$ & $(\frac{1}{n}\sum_{j=1}^{n}\mathbf{Q}_{i,j}^{t})\oplus \mathbf{p}_i^{t}$ & $\mathbf{p}_i^t$\\\hline
 
 $\mathbf{p}_i^t$, $\mathbf{S}_i^t$ & $\rm{GRU}( \mathbf{S}_i^t, \mathbf{p}_i^t )$, feature dimension: $64$ & $\mathbf{S}_i^t$ \\\hline
 
$\mathbf{S}_i^t$ & $64 \to 3$  & \\\hline
\end{tabular}
\label{tab:AIM_decoder}
\end{table}
$\rm{GRU}(\cdot)$ denotes a GRU unit, whose hidden feature dimension is $64$. It predicts the future position of all joints conditioned on A-links and previous frames.

\subsection*{Backbone}
The backbone network of  the AS-GCN extracts the rich spatial and temporal feature of actions with the proposed ASGC and temporal CNN (T-CN). For example, we build AS-GCN on NTU-RGB+D dataset and Cross-Subject benchmark~\cite{Shahroudy_2016_CVPR}. There are 25 joints, 3D spatial positions and 300 padded frames for each action. The architecture of the backbone is presented in Table~\ref{tab:backbone}.
\begin{table}[h]
\small
\centering
\caption{\small The architecture of the backbone network of AS-GCN}

\begin{tabular}{|c|c|c|}
\hline
In-Shape & Operation Shape & Out-Shape \\
\hline\hline
\multirow{2}{*}{[25,3,300]} &  ASGC:[64,1,64,1]$\times$($n_A$+$n_S$) & \multirow{2}{*}{[25,64,300]} \\
                            & T-CN:[64,1,64,7], stride=1&                              \\\hline
\multirow{2}{*}{[25,64,300]} & ASGC:[64,1,64,1]$\times$($n_A$+$n_S$) & \multirow{2}{*}{[25,64,300]} \\
                            & T-CN:[64,1,64,7], stride=1&                              \\\hline
\multirow{2}{*}{[25,64,300]} & ASGC:[64,1,64,1]$\times$($n_A$+$n_S$) & \multirow{2}{*}{[25,64,300]} \\
                            & T-CN:[64,1,64,7], stride=1&                              \\
\hline\hline
\multirow{2}{*}{[25,64,300]} & ASGC:[64,1,64,1]$\times$($n_A$+$n_S$) & \multirow{2}{*}{[25,128,150]} \\
                            & T-CN:[128,1,64,7], stride=2&                              \\\hline
\multirow{2}{*}{[25,128,150]} & ASGC:[128,1,128,1]$\times$($n_A$+$n_S$) & \multirow{2}{*}{[25,128,150]} \\
                            & T-CN:[128,1,128,7], stride=1&                              \\\hline
\multirow{2}{*}{[25,128,150]} & ASGC:[128,1,128,1]$\times$($n_A$+$n_S$) & \multirow{2}{*}{[25,128,150]} \\
                            & T-CN:[128,1,128,7], stride=1&                              \\
\hline\hline
\multirow{2}{*}{[25,128,150]} & ASGC:[128,1,128,1]$\times$($n_A$+$n_S$) & \multirow{2}{*}{[25,256,75]} \\
                            & T-CN:[256,1,128,7], stride=2&                              \\\hline
\multirow{2}{*}{[25,256,75]} & ASGC:[256,1,256,1]$\times$($n_A$+$n_S$) & \multirow{2}{*}{[25,256,75]} \\
                            & T-CN:[256,1,256,7], stride=1&                              \\\hline
\multirow{2}{*}{[25,256,75]} & ASGC:[256,1,256,1]$\times$($n_A$+$n_S$) & \multirow{2}{*}{[25,256,75]} \\
                            & T-CN:[256,1,256,7], stride=1&                              \\
\hline
\end{tabular}

\label{tab:backbone}
\end{table}
There are nine ASGC blocks consisting of the backbone of AS-GCN model. The input/output feature maps are 3D tensors, where the three axes represent the joint number, feature dimension and frame number, respectively. The shapes of operations have the consistent dimensions with input and output feature maps, where the first axis is the filter number or output feature dimension, and the other three correspond to the input shape. $A$ and $B$ are the types of A-links and S-links. For action recognition, we obtain the last feature map whose shape is [25,256,75] and apply a global average pooling operation on the time and joint axis, i.e. the 1st and 3rd axis. Thus we obtain a semantic feature vector of the action, whose dimension is 256.

\subsection*{Future Action Prediction Head}
The architecture of the future action prediction head of AS-GCN model are presented in Table~\ref{tab:prediction}.
\begin{table}[h]
\small
\centering
\caption{\small The architecture of the prediction head of AS-GCN model}

\begin{tabular}{|c|c|c|}
\hline
In-Shape & Operation Shape & Out-Shape \\
\hline\hline
\multirow{2}{*}{[25,256,75]} & ASGC:[128,1,256,1]$\times$($n_A$+$n_S$) & \multirow{2}{*}{[25,128,39]} \\
                            & T-CN:[128,1,128,7], stride=2&                              \\\hline
\multirow{2}{*}{[25,128,39]} & ASGC:[128,1,128,1]$\times$($n_A$+$n_S$) & \multirow{2}{*}{[25,128,19]} \\
                            & T-CN:[128,1,128,7], stride=2&                              \\\hline
\multirow{2}{*}{[25,128,19]} & ASGC:[128,1,128,1]$\times$($n_A$+$n_S$) & \multirow{2}{*}{[25,128,10]} \\
                            & T-CN:[128,1,128,7], stride=2&                              \\\hline
\multirow{2}{*}{[25,128,10]} & ASGC:[128,1,128,1]$\times$($n_A$+$n_S$) & \multirow{2}{*}{[25,128,5]} \\
                            & T-CN:[128,1,128,3], stride=2&                              \\\hline
\multirow{2}{*}{[25,128,5]} & ASGC:[128,1,128,1]$\times$($n_A$+$n_S$) & \multirow{2}{*}{[25,128,1]} \\
                            & T-CN:[128,1,128,5], stride=1&                              \\\hline
\multirow{2}{*}{[25,131,1]} & ASGC:[64,1,131,1]$\times$($n_A$+$n_S$) & \multirow{2}{*}{[25,64,1]} \\
                            & T-CN:[64,1,64,1], stride=1&                              \\\hline
\multirow{2}{*}{[25,67,1]} & ASGC:[32,1,67,1]$\times$($n_A$+$n_S$) & \multirow{2}{*}{[25,32,1]} \\
                            & T-CN:[32,1,32,1], stride=1&                              \\\hline
\multirow{2}{*}{[25,35,1]} & ASGC:[30,1,35,1]$\times$($n_A$+$n_S$) & \multirow{2}{*}{[25,30,1]} \\
                            & T-CN:[30,1,30,1], stride=1&                              \\\hline
{[25,33,1]} & FC:[30,1,33,1] &{[25,30,1]} \\\hline
\end{tabular}

\label{tab:prediction}
\end{table}
We input the output feature map from the backbone network in to the prediction head. The input tensor are calculated by nine ASGC blocks. The first five blocks reduce the frame number to aggregate higher-level action features. The last four blocks work on action regeneration. For the last four blocks, we concatenate the last input frame to each feature map. Finally, with a residual connection, we obtain a tensor with shape [30,1,25] from a fully connected layer, which contains the joint position of the predicted 10 frames.

\section*{Appendix C: More Future Pose Predictions}
\begin{figure*}[h]
  \centering
  \subfloat[One action of 'wipe face'. The 117th to 126th frames are predicted from the 116th frame.]{\includegraphics[width=0.98\textwidth]{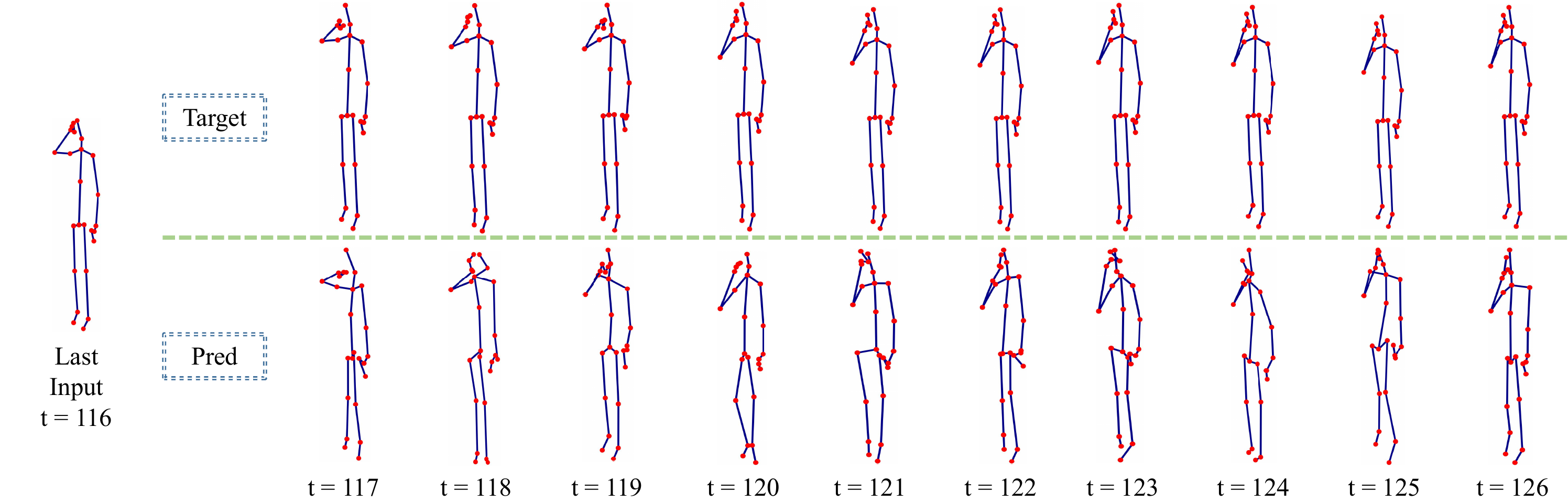}} \\
  \subfloat[One action of 'throw'. The 69th to 78th frames are predicted from the 68th frame.]{\includegraphics[width=0.98\textwidth]{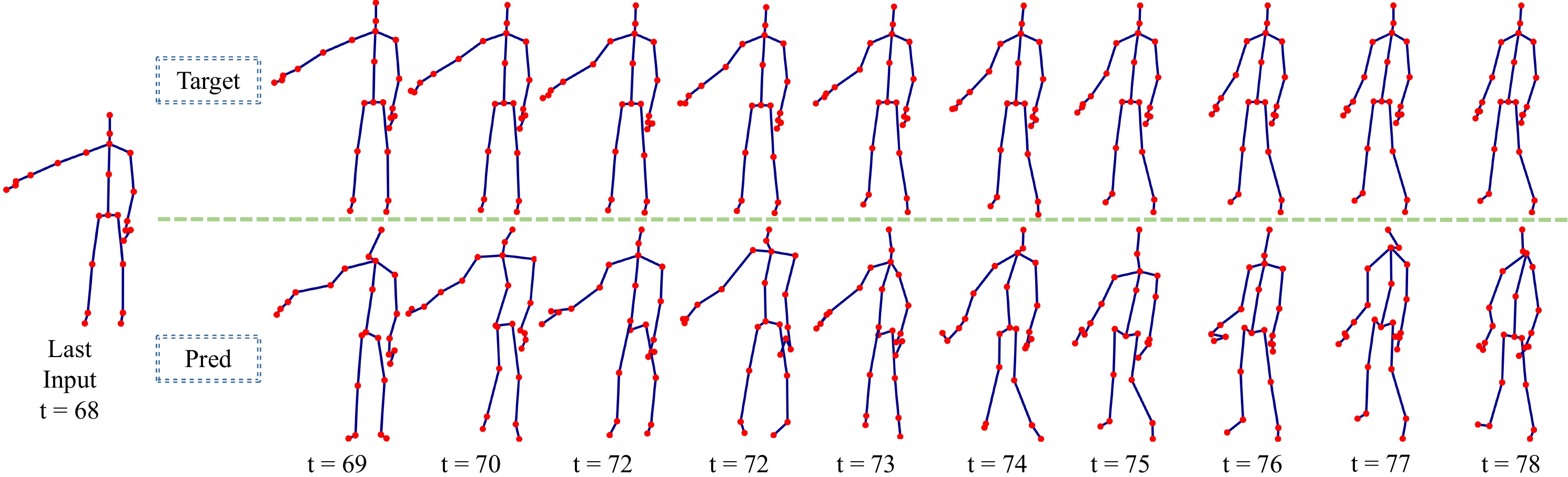}} \\
  \subfloat[One action of 'nausea or vomiting condition'. The 91th to 100th frames are predicted from the 90th frame.]{\includegraphics[width=0.98\textwidth]{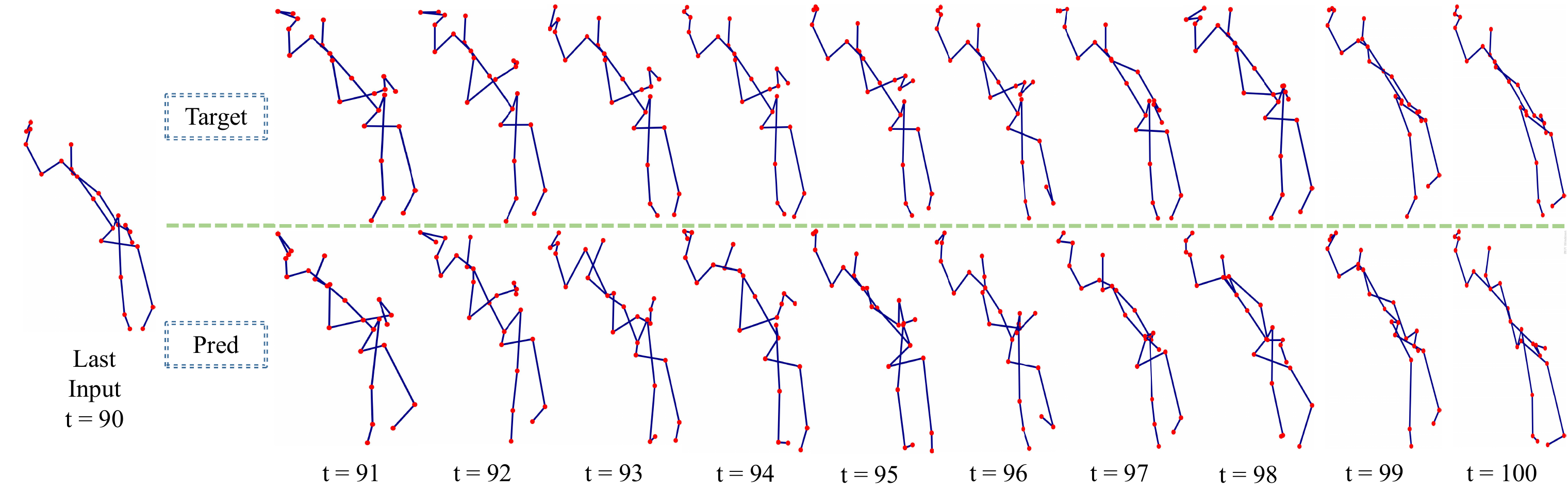}} 
  \caption{\small Some predicted actions, including 'wipe face', 'throw' and 'nausea or vomiting codition' in NTU-RGB+D dataset.}
	\label{fig:geo_distribution}
	\vspace{0.2in}
\end{figure*}

More future action prediction results of different actions are illustrated in Figure~\ref{fig:geo_distribution},
 which contains the action of 'wipe face', 'throw' and 'nausea or vomiting condition'.
As we see, the actions are predicted with very low error. 


\end{document}